\documentclass{article}

     \PassOptionsToPackage{numbers, compress, sort}{natbib}

\usepackage[preprint]{neurips_2021}

\usepackage[utf8]{inputenc} %
\usepackage[T1]{fontenc}    %
\usepackage{hyperref}       %
\usepackage{url}            %
\usepackage{booktabs}       %
\usepackage{amsfonts}       %
\usepackage{nicefrac}       %
\usepackage{microtype}      %
\usepackage{xcolor}         %
\usepackage{wrapfig}
\usepackage{amsmath}
\usepackage{cleveref}
\usepackage{csquotes}
\usepackage{graphicx}
\usepackage{multirow}
\usepackage{enumitem}
\usepackage{xspace}
\usepackage{multicol}
\usepackage{tabu}
\usepackage{xcolor}

\usepackage{amsmath,amsfonts,bm}

\def\eqref#1{equation~\ref{#1}}

\def\1{\bm{1}}

\def\vy{{\bm{y}}}

\def\mX{{\bm{X}}}
\def\mY{{\bm{Y}}}

\DeclareMathAlphabet{\mathsfit}{\encodingdefault}{\sfdefault}{m}{sl}
\SetMathAlphabet{\mathsfit}{bold}{\encodingdefault}{\sfdefault}{bx}{n}

\definecolor{bleudefrance}{rgb}{0.19, 0.55, 0.91}
\definecolor{cornflowerblue}{rgb}{0.39, 0.58, 0.93}
\definecolor{darkcerulean}{rgb}{0.03, 0.27, 0.49}
\definecolor{deepskyblue}{rgb}{0.0, 0.75, 1.0}

\definecolor{deepgreen}{rgb}{0.0, 0.6, 0.0}

\newcommand{\eat}[1]{}

\newcommand{\paragraphsq}[1]{\vspace{-5pt}\paragraph{#1}}

\makeatletter
\g@addto@macro\small{%
  \setlength\abovedisplayskip{-5pt}
  \setlength\abovedisplayshortskip{-5pt}
  \setlength\belowdisplayshortskip{-5pt}
  \setlength\belowdisplayskip{-5pt}
}
\makeatother

\newcommand{\Methodfull}{Pseudo Subword\xspace}
\newcommand{\methodfull}{pseudo subword\xspace}

\newcommand{\methodfulls}{pseudo subwords\xspace}
\newcommand{\modelfull}{Wav2Seq\xspace}
\newcommand{\modelfulls}{Wav2Seqs\xspace}
\newcommand{\Modelfull}{Wav2Seq\xspace}

\newcommand{\modelsize}[1]{Wav2Seq (from #1)\xspace}

\newlength\savewidth
\newcommand{\tablestyle}[2]{\setlength{\tabcolsep}{#1}\renewcommand{\arraystretch}{#2}\centering\footnotesize}
\makeatletter\renewcommand\paragraph{\@startsection{paragraph}{4}{\z@}
  {.5em \@plus1ex \@minus.2ex}{-.5em}{\normalfont\normalsize\bfseries}}\makeatother

\newcommand*{\shortautoref}[1]{%
  \begingroup
    \def\sectionautorefname{Sec.}%
    \def\subsectionautorefname{Subsec.}%
    \def\subsubsectionautorefname{Subsubsec.}%
    \def\figureautorefname{Fig.}%
    \autoref{#1}%
  \endgroup
}

\title{Wav2Seq: Pre-training Speech-to-Text Encoder-Decoder Models Using Pseudo Languages}

\author{
  Felix Wu$\phantom{}^{\dagger}$  \hspace{2pt} 
  Kwangyoun Kim$\phantom{}^{\dagger}$  \hspace{2pt} 
  Shinji Watanabe$\phantom{}^{\diamond}$ \hspace{2pt}
  Kyu Han$\phantom{}^{\dagger}$  \hspace{0pt} 
  \\[3pt]
  \textbf{Ryan McDonald$\phantom{}^{\dagger}$  \hspace{2pt} 
  Kilian Q. Weinberger$\phantom{}^{\dagger\ddagger}$ \hspace{2pt}
  Yoav Artzi$\phantom{}^{\dagger\ddagger}$}
  \\[6pt]
  $\phantom{}^{\dagger}$ASAPP Inc.\hspace{15pt} 
  $\phantom{}^{\diamond}$Carnegie Mellon University  \hspace{15pt}
  $\phantom{}^{\ddagger}$Cornell University 
  \\
  \texttt{\{fwu, kkim, khan, rmcdonald, kweinberger, yoav\}@asapp.com} \\
}

\begin{document}

\maketitle

\begin{abstract}
We introduce Wav2Seq, the first self-supervised approach to pre-train \emph{both parts} of encoder-decoder models for speech data.
We induce a pseudo language as a compact discrete representation, and formulate a self-supervised pseudo speech recognition task --- transcribing audio inputs into pseudo subword sequences.
This process stands on its own, or can  be applied as low-cost second-stage pre-training. 
We experiment with automatic speech recognition (ASR), spoken named entity recognition, and speech-to-text translation.
We set new state-of-the-art results for end-to-end spoken named entity recognition, and show consistent improvements on 20 language pairs for speech-to-text translation, even when competing methods use additional text data for training. On ASR, our approach enables encoder-decoder methods to benefit from pre-training for all parts of the network, and shows comparable performance to highly optimized recent methods.

\end{abstract}

\section{Introduction}
\label{sec:intro}

Self-supervised pre-trained models have recently become a core part of speech models~\citep{Oord2018RepresentationLW,Schneider2019wav2vecUP,baevski2020wav2vec2,hsu2020hubert,chen2021wavlm,wu2021performance,chung2021w2v,grill2020bootstrap}, leading to impressive performance on a wide variety of speech tasks~\citep{Panayotov2015LibrispeechAA,wang2020covost2,yang2021superb,shon2021slue}. 
This trend mirrors recent success in natural language processing~\citep[NLP;][]{peters2018deep,Devlin2019BERT,liu2019roberta,radford2019language,lewis2020bart,raffel2019exploring} and computer vision~\citep[CV;][]{wu2018unsupervised,chen2020simple,he2020momentum,caron2021emerging,li2021efficient}.

Most of these approaches rely on pre-training an encoder to create expressive representation of input data. If a sequential decoder is needed for downstream tasks (i.e., for generative tasks), it is often trained with task-specific supervised data. 
The most common approaches for automatic speech recognition (ASR) follow this encoder-decoder paradigm~\citep{Collobert2016Wav2LetterAE,watanabe2018espnet,pratap2019wav2letter++,Han2020ContextNetIC,Gulati2020ConformerCT,baevski2022data2vec}, regardless if they use sequence transducers~\citep{graves2012sequence} or sequence-to-sequence~\citep{sutskever2014sequence,chorowski2015attention,chan2016listen} architectures.
Because all existing self-supervised learning approaches for speech focus on pre-training an encoder model only, when adapted to an encoder-decoder architecture, the decoder has to be either randomly initialized or borrowed from a pre-trained NLP decoder~\citep{Wang2021LargeScaleSA,babu2021xls}.

\begin{figure}[t]
    \centering
    \includegraphics[width=0.8\linewidth]{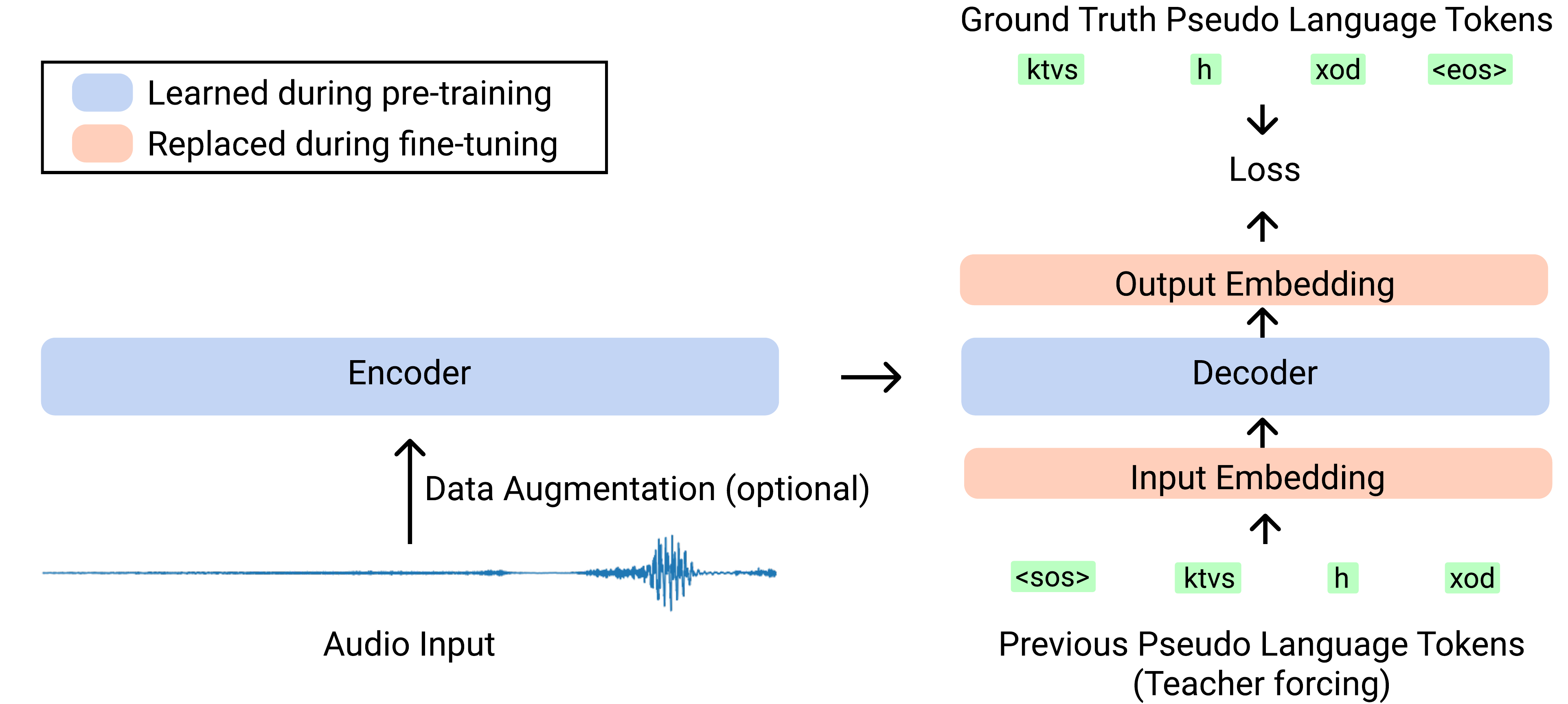}
    \caption{
    Pseudo ASR task. The \modelfull model is pre-trained to transcribe an audio input into a sequence of pseudo language tokens.
    The embedding layers are replaced during fine-tuning on real ASR or speech translation tasks. ``$<$sos$>$'' is the start-of-sequence token and ``$<$eos$>$'' is the end-of-sequence token.
    See \shortautoref{fig:pseudo_language_creation} for how the pseudo language tokens are generated.
    }
    \label{fig:pseudo-asr}
\end{figure}

In this paper, we propose \modelfull, the first self-supervised approach to jointly pre-train the encoder \emph{and} decoder. 
We automatically induce \emph{\methodfulls} that form a compact discrete representation of spoken language.
We treat these as audio transcripts in a pseudo ASR task, and use them as the targets for Seq2Seq learning  (see \shortautoref{fig:pseudo-asr}).
When fine-tuned on a downstream task (e.g., ASR or speech translation), the input and output embedding layers are replaced in order to adapt to natural language.

We conduct extensive experiments on ASR, spoken named entity recognition (SNER), and speech-to-text translation (ST) tasks. 
Focusing on settings with limited labeled audio data (i.e., 10h or less), our ASR results show that with a pre-trained encoder only, CTC models outperform encoder-decoder models significantly in few-example scenarios; however, \modelfull boosts the performance of encoder-decoder models and closes this gap.
When we apply \modelfull as a low-cost second-stage pre-training method, models based on existing pre-trained encoders (e.g., HuBERT) achieve even better results.
On the SLUE-VoxPopuli SNER benchmark, \modelfull initialized with HuBERT achieves the best end-to-end results. For ST tasks, we conduct experiments on four from-English language pairs and 16 to-English pairs with both low- and high-resource setups. \Modelfull consistently outperforms models initialized with  HuBERT or XLS-R pre-trained models and achieves similar BLEU scores as models trained on additional machine translation annotated text data. 
Pre-trained models and code are available at \url{https://github.com/asappresearch/wav2seq}.

\section{Related Work}\label{sec:related_work}

\paragraph{Speech-to-Text Models}
Automatic speech recognition (ASR)  and speech-to-text translation (ST) are two of the most commonly studied speech-to-text tasks.
The former produces a monotonic mapping while the latter may involve re-ordering (e.g. translating English to German).
When re-ordering exists, it is advantageous to use  sequence-to-sequence (Seq2Seq) models~\citep{sutskever2014sequence} with encoder-decoder attention~\citep{bahdanau2014neural,chan2016listen}, such as doen in popular open-source speech translation toolkits~\citep{wang2020fairseqs2t,inaguma2020espnet}.

When the mapping is monotonic, such as with ASR, various approaches are applicable. 
Connectionist temporal classification \citep[CTC;][]{Graves2006ctc} is one of the simplest and efficient approaches.
It only requires an encoder that generates a sequence of feature vectors where each feature vector represents the input within a time window.
A linear classifier can be applied to classify each feature vector into a character and the repeated and blank characters will be removed.
When decoding with an LM, beam-search can also be used.
CTC is used in various systems~\citep{miao2015eesen,amodei2016deep,soltau2016neural} and is becoming the default approach to fine-tune a self-supervised pre-trained model such as wav2vec 2.0~\citep{baevski2020wav2vec2}.
The sequence transducer~\citep{graves2012sequence} is another architecture dedicated to monotonic mapping in ASR.
Similar to Seq2Seq, it has an  encoder for audio inputs and an auto-regressive decoder for text inputs; however, the decoder does not have access to the full encoded speech features. 
The decoder starts with using the speech feature of the first frame from the encoder and decides either to emit a text token or shift to use the speech feature of the next frame.
State-of-the-art supervised speech models are based on the transducer framework~\citep{Han2020ContextNetIC,Gulati2020ConformerCT}.

\paragraph{Text-to-Text Encoder-Decoder Pre-training}
In NLP, it has been observed that pre-trained Seq2Seq models outperforms pre-trained encoders on text generation tasks~\citep{lewis2020bart,raffel2019exploring} despite being less competitive in discriminative tasks.
\citet{lewis2020bart} introduced BART which is a Seq2Seq Transformer pre-trained on a text denoising task. 
\citet{Xu2020SelftrainingAP} extend BART to a multilingual setup and introduce mBART.
Concurrently, T5~\citep{raffel2019exploring} is pre-trained on a large collection of NLP tasks, showing effective task transfer behavior~\citep{zhang2020pegasus,sanh2021multitask,qi2020prophetnet}.

\paragraph{Speech Self-supervised Learning}
CPC~\citep{Oord2018RepresentationLW} and wav2vec \citep{Schneider2019wav2vecUP} are two early approaches for self-supervised speech representation learning  based on contrastive loss.
Wav2vec 2.0~\citep{baevski2020wav2vec2} is the first self-supervised model to outperform purely supervised approaches on ASR.
Instead of directly reconstructing masked features, it is uses a contrastive loss --- distinguishing the quantized version of the correct features from several negative samples.
\citet{lai2021parp} show that many of the weights in wav2vec 2.0 are redundant and can be pruned. 
\citet{wu2021performance} provides a deeper understanding of the efficiency of various components of wav2vec 2.0 and propose SEW-D, a more compact and efficient variant.
\citet{Hsu2021RobustW2} found that adding a loss to intermediate layer improves the quality of wav2vec 2.0 pre-training.
Rather than contrastive learning, \citet{hsu2020hubert} propose HuBERT, a self-supervised approach based on masked language modeling.
Unlike wav2vec 2.0 using online quantization, HuBERT extracts speech features and clusters them offline. The model is trained to predict the cluster indices of the features at the masked positions.
WavLM~\citep{chen2021wavlm} extends HuBERT by using a larger dataset with non-audiobook audios and adding noise to the audio during pre-training. 
W2v-BERT~\citep{chung2021w2v} further combines these two approaches (contrastive learning and masked language modeling) and sets a new state of the art in ASR. 
Concurrently to our paper, \citet{baevski2022data2vec} introduce data2vec which is trained to match the outputs of an exponentially moving averaged teacher (similar to BYOL~\citep{grill2020bootstrap} and DINO~\citep{caron2021emerging} in CV) and works for NLP and CV data as well.
\citet{ao2021speecht5} introduce SpeechT5, a multimodal encoder-decoder model, which is pre-trained on both text and speech data. Unlike our approach, their decoder is trained to reconstruct the log Mel-filterbank features of the speech audio.

\section{Self-supervision using {\Methodfull}s} \label{sec:method}
\subsection{Pseudo ASR}

Seq2seq ASR models use an encoder to extract speech features and a decoder to generate text tokens conditioned on the speech features.
Given an audio input $\mX = (x_1, x_2, ..., x_m) \in \mathcal{X}$, the ASR model $f: \mathcal{X} \rightarrow \mathcal{Y}$ is trained to transcribe it into text tokens $\mY = (y_1,  y_2, .., y_n) \in \mathcal{Y}$, where $\mathcal{X}$ is the space of audio and $\mathcal{Y}$ is the space of text sentences and $m$ and $n$ are input and output lengths.
Seq2seq models are commonly trained by minimizing the negative log likelihood using training data $\mathcal{D}_{\text{train}}$ of audio-transcription pairs:
\begin{equation}
\ell_{\text{NLL}} = - \sum_{(\mX, \mY) \in \mathcal{D}_{\text{train}}} 
\sum_{t=1}^{n} 
\log P(\vy_t | \mX, y_1, ..., y_{t-1})\;\;.
\end{equation}
 During pre-training, in contrast, transcriptions are not available, making this objective inapplicable.  
Our key insight is that we can use unsupervised techniques to induce a pseudo language given raw audio only, and annotate raw audio inputs with it automatically. We then apply the likelihood sequence-to-sequence objective, and train \Modelfull, a pre-trained encoder-decoder architecture.   
\shortautoref{fig:pseudo-asr} illustrates the pseudo ASR task.

While our focus is the negative log-likelihood objective, the pseudo language formulation is not restricted to this objective, and is broadly compatible with other training objectives. 
For example, we also experiment with pre-training \modelfulls with a sequence transducer architecture using a transducer loss~\citep{graves2012sequence} (\shortautoref{subsubsec:small_model_exp}).

\subsection{Inducing Pseudo Languages}

We construct sequences of discrete tokens from speech. Because Transformer decoders require quadratic computation with respect to sequence length, we need  to balance the length of these sequences (i.e., wanting them to be short) with the quality of the pre-trained model. 

\begin{wrapfigure}{r}{0.5\textwidth}
  \begin{center}
    \includegraphics[width=0.48\textwidth]{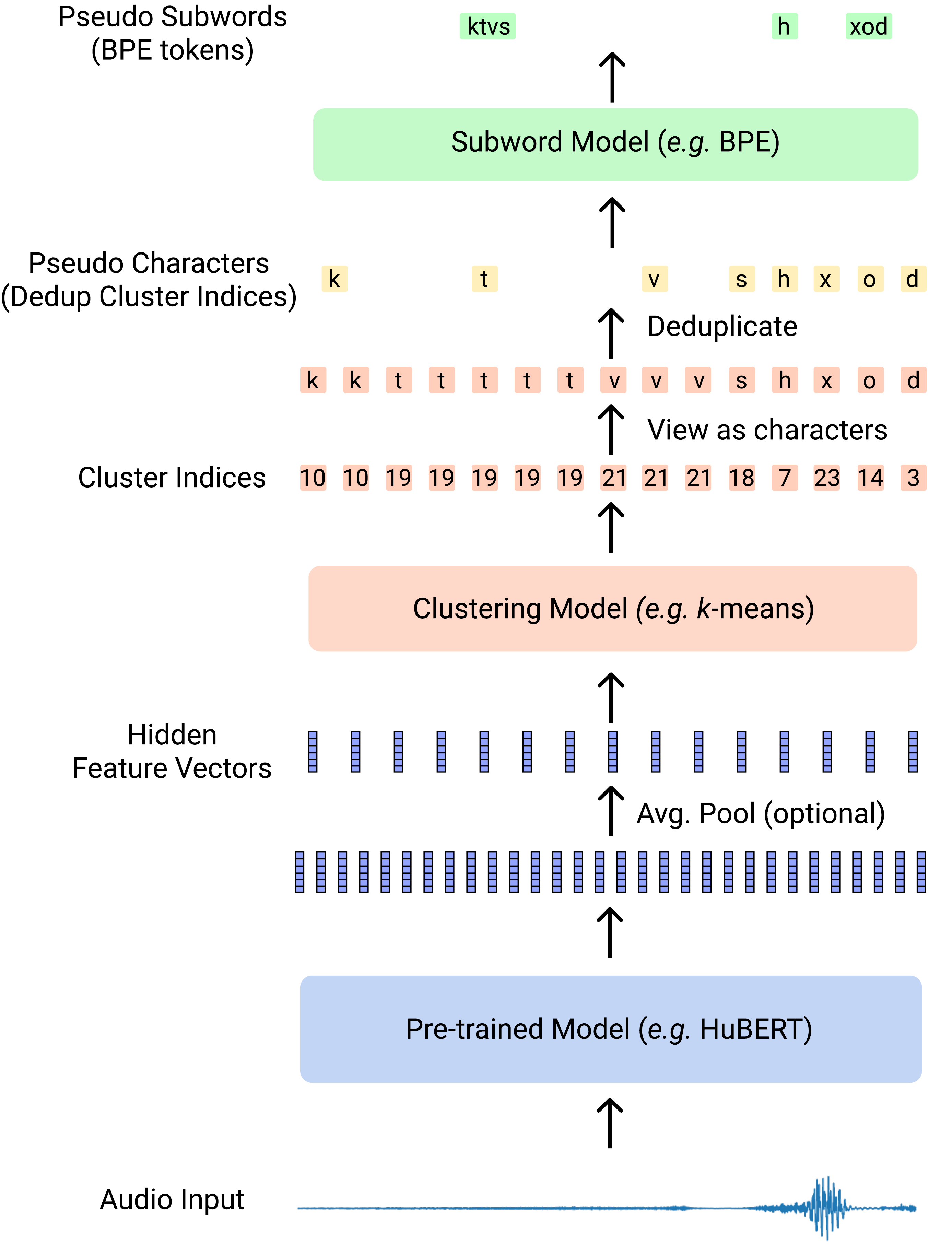}
  \end{center}
   \caption{Mapping an audio input into \methodfulls. Here we use the first 0.6 seconds of a real audio from LibriSpeech~\citep{Panayotov2015LibrispeechAA} training set as an example.
    We use models with 25 clusters and 1000 BPE tokens in this example. 
    Pseudo subword is a compact representation (in terms of the sequence length) of the audio input.}
    \label{fig:pseudo_language_creation}
\end{wrapfigure}

\shortautoref{fig:pseudo_language_creation} illustrates how our {\methodfull}s are generated.
We extract a sequence of hidden feature vectors from an audio file using a pre-trained HuBERT~\citep{hsu2020hubert} model. We apply average pooling to reduce the sequence length and then \textit{k}-means clustering to discretize these hidden feature vectors. This results in a sequence of cluster indices, which have also been referred to as \emph{hidden units} in prior literature~\cite{hsu2020hubert}.

We treat these cluster indices as \emph{characters} because they have been found to correlate with the phonemes of speech audio.
We observe many consecutive repetitions of cluster indices, and propose to deduplicate them (i.e., remove the duplicates). For example, ``k k t t t'' will be mapped to ``k t''. 
We treat the deduplicated cluster indices as the characters of \methodfull{}s and refer to them as \emph{pseudo characters}.

We apply a subword tokenization algorithm such as byte-pair encoding~\citep[BPE;][]{gage1994new} to further shorten the target sequence length.
This process merges common co-occurring characters into newly created tokens, and is widely used in NLP tasks such as language modeling and machine translation to balance between the benefits and costs of character and word representations (i.e., vocabulary size, unit semantics, handling of unseen words, etc). 
For example, a word ``negotiation`' can be represented by ``ne'' ``go'' ``ti'' and ``ation'' with a subword tokenizer, which allows the model to share the embedding of the common suffix ``ation'' across different words. 
Similarly, common pseudo characters sequences are merged to \emph{pseudo subwords}.

\section{Experimental Setup}
\subsection{Data}
\paragraphsq{LibriSpeech}
We use LibriSpeech~\citep{Panayotov2015LibrispeechAA} data for pre-training and focus on low-resource ASR using 10h and 100h subsets of labels.
LibriSpeech contains 960 hours of audiobook recordings for training, two 5-hour development sets (dev-clean and dev-other) and two 5-hour test sets (test-clean and test-other).
For self-supervised pre-training, we randomly sample 1\% of the 960h training set as the validation set and use the rest of the raw data (i.e., without labels) as the training set.
For ASR experiments, we use the LibriSpeech train-clean 100h split and the 10h splits provided by \citet{librilight} as the low-resource labelled data. We use dev-other for hyper-parameter tuning, ablations, and early stopping.

\paragraphsq{LibriLight}
Similar to the training of wav2vec 2.0 large and HuBERT-large, we use LibriLight~\citep{librilight} to pre-train large  models. LibriLight contains 60K hours of unlabelled audiobook recordings.

\paragraphsq{CoVoST-2}
We use CoVoST-2 speech translation data set~\citep{wang2020covost2} for ST experiments. There are 21 X-to-English and 15 English-to-X language pairs. For X-to-English pairs, we use the 12 low-resource pairs with less than 10h audios and 4 high-resource pairs with more than 100h audios.
All English-to-X pairs have 430h of English audio.  
We therefore use the four language pairs used in prior work~\citep{Wang2021LargeScaleSA} and subsample a 10h subset for each pair to simulate a low-resource setup.

\paragraphsq{SLUE-VoxPopuli}
We use SLUE-VoxPopuli dataset~\citep{shon2021slue} for spoken NER experiments. SLUE-VoxPopuli contains several subsets: fine-tune, dev, and test sets with 14.5h, 5h, and 5h of transcribed audios and their corresponding named entity labels. The audio is from recordings of  European Parliament events and the transcriptions are provided by \citet{Wang2021VoxPopuliAL}.

\subsection{Pre-training}
We use the official baselines for most prior work~\citep{baevski2020wav2vec2,hsu2020hubert,babu2021xls}, which are implemented in fairseq~\citep{Ott2019fairseqAF}. 
Our \modelfull is also implemented as a plugin of fairseq. We use the mini-batch k-means algorithm~\citep{sculley2010web} with k-mean++ initialization~\citep{k-mean++} implementation in scikit-learn~\citep{scikit-learn}. 
Following HuBERT's best hyper-parameters, the number of clusters $C$ is set to 500 by default unless specified separately.
We use the byte-pair encoding (BPE)~\citep{gage1994new} implementation from Huggingface's tokenizers library\footnote{\url{https://github.com/huggingface/tokenizers}} as the subtokenization method where the vocabulary size of subword tokens $V$ is set to 30K (or 10K for the tiny models).

Following \citet{baevski2020wav2vec2}, we pre-trained models with a batch size of at most 87.5 seconds per GPU on 8 GPUs and an update frequency of 4 to simulate 32-GPU training as \citet{baevski2020wav2vec2}.
We pre-train the models for 100K updates (or 25K updates in second-stage pre-training) unless specified otherwise.
We use the same hyper-parameters as prior work~\citep{baevski2020wav2vec2, wu2021performance}. The peak learning rate is set to $2 \times 10^{-4}$ with 32K linear warm-up steps and linearly decaying to 0 at step 400K.
All models are pre-trained with eight NVIDIA V100 GPUs with half-precision. 
We tie the weights of input and output embeddings of the decoder during pre-training and fine-tuning~\citep{inan2016tying,press2017using}.

\subsection{Fine-tuning}
We use BPE with vocabulary size 1{,}000 to tokenize the text in ASR, spoken NER, or ST tasks.
Similar to prior work~\citep{baevski2020wav2vec2,hsu2020hubert}, we use learning rate $5 \times 10^{-5}$ for fine-tuning and use tri-stage learning rate scheduler with 2{,}000 steps for linear warm-up and the last 50\% of the updates for exponential decay.\footnote{We tune the learning rate $\in \{2 \times 10^{-4}, 10^{-4}, 5 \times 10^{-5}, 2\times 10^{-5}\}$ on the small models on ASR and fix them.} 
The number of fine-tuning steps depends on the datasets. When fine-tuned on LibriSpeech 10h (or 100h) data, we use a batch size with at most 50 seconds per GPU of audio for 20K (or 80K) and updates on eight GPUs with half-precision. For spoken NER and low-resource ST, we use the same hyper-parameters as the LibriSpeech 10h setup. 
For high resource ST, we increase the number of updates to 320K because it has more data. We use beam size 10 for beam search decoding of Seq2Seq models and no length penalty for ASR and SNER tasks. For ST, we tune length penalty $\in \{0.5, 1.0, 1.5\}$ on the development sets because some languages may be more (or less) compact. For CTC baselines, we follow the hyperparamters provided by \citet{hsu2020hubert}.
\section{Experiments}

\subsection{Automatic Speech Recognition (ASR)}

\subsubsection{Small Model Experiment} \label{subsubsec:small_model_exp}

We conduct initial small-scale experiments using tiny models with embedding size 256, four attention heads, and feed-forward embedding size 1{,}024 in each Transformer block. Each model has 6 or 12 Transformer blocks. We use a compact wave feature extractor (WFE-C-c128l0) \cite{wu2021performance} to speed up the model. This extractor has been shown to perform similar to the wave feature extractor used in wav2vec 2.0 and HuBERT, but faster.
All models are pre-trained on LibriSpeech with a semi-supervised setup using 960h unlabelled recordings for pre-training and 10h labelled data for fine-tuning.
To be comparable, we use the 9th layer of the official second iteration HuBERT-base model to extract both hidden units and \methodfulls for pre-training HuBERT baselines and our \modelfull models. All HuBERT models in this subsubsection are third-iteration of HuBERT models.

\begin{table}[t]
    \tablestyle{2pt}{1.0}
    \scriptsize
    \centering
    \begin{tabu}{cclcccc}
    \toprule
    \multirow{2}{*}{ ASR Type } & \multirow{2}{*}{\#} & \multirow{2}{*}{ Pre-training Method } & Pre-train  & \multicolumn{2}{c}{Layers} & dev-other \\
    & & & Iterations & Enc & Dec & WER (\%) \\
    \midrule
    \multirow{2}{*}{ CTC }
    & 1 & HuBERT & 100K & 6 & 0 & 51.7 \\
    & 2 & HuBERT & 100K & 12 & 0 & 36.5 \\
    \midrule
    \multirow{6}{*}{ Seq2Seq }
    & 3 & No pre-training & 0 & 6 & 6 & $\ge$100.0 \\
    & 4 & HuBERT & 100K & 6 & 6 & $\ge$100.0 \\
    & 5 & HuBERT & 100K & 12 & 6 & 85.8 \\
    \rowfont{\color{blue}}
    & 6 & \modelfull & 100K & 6 & 6 & 38.1 \\
    \rowfont{\color{blue}}
    & 7 & \modelfull & 100K & 8 & 4 & 36.6 \\
    \rowfont{\color{blue}}
    & 8 & \modelsize{HuBERT} & 100K + 25K & 6 & 6 & 34.7 \\
    \midrule
    \multirow{4}{*}{ Transducer }
    & 9 & No pre-training & 0 & 6 & 3 & 93.5 \\
    & 10 & HuBERT & 100K & 6 & 3 & 93.6 \\
    \rowfont{\color{blue}}
    & 11 & \modelfull & 100K & 6 & 3 & 44.2 \\
    \rowfont{\color{blue}}
    & 12 & \modelsize{HuBERT} & 100K + 25K & 6 & 3 & 40.8 \\
    \bottomrule
    \end{tabu}
    \caption{Small model ASR experiment with unsupervised pretraining on LibriSpeech 960h and fine-tuning with LibriSpeech 10h ASR labels. All models have embedding size 256, FFN size 1{,}024, and four attention heads in their Transformer layers. We show that \modelfull works with both Seq2Seq and Transducer architectures. Moreover, \modelfull is complementary to HuBERT pre-training --- when initializing the encoder with HuBERT and having a second stage pre-training with \modelfull, we observe the best results.}
    \label{tab:ls-10h-small}
\end{table}
\begin{table}[t]
    \tablestyle{3pt}{1.0}
    \scriptsize
    \centering
    \begin{tabu}{c|l|cc|cccc}
    \toprule
    Labels & Model & PT Iter. & ASR Iter. & dev-c & dev-o & test-c & test-o \\
    \midrule
    10h & S2T & 0 & 15K & $\ge$100  & $\ge$100 & $\ge$100  & $\ge$100  \\
    \rowfont{\color{blue}}
    10h & \modelfull-S2T & 300K & 5K & 45.0 & 54.5 & 48.0 & 55.1  \\
    \midrule
    100h & S2T & 0 & 150K & 16.4 & 31.6 & 16.9 & 32.8 \\
    \rowfont{\color{blue}}
    100h & \modelfull-S2T & 300K & 50K & 12.6 & 25.8 & 12.8 & 26.8 \\
    \bottomrule
    \end{tabu}
    \caption{LibriSpeech WER (\%) using fairseq S2T Transformer-small model~\citep{wang2020fairseqs2t}. 10h or 100h of labelled transcripts are provided. With 10h labelled data, the model cannot learn effectively without \modelfull pre-training. With 100h labelled data, pre-training the S2T model using \modelfull on 960h unlabelled audio results in the same model architecture achieving better performance with only 1/3 of the training time.}
    \label{tab:s2t-exp}
    \vspace{-10pt}
\end{table}

\shortautoref{tab:ls-10h-small} shows word error rate (WER) of HuBERT and \modelfull with different fine-tuning setups.
In the first two rows, as is already known, HuBERT performs strongly when fine-tuning with CTC objective which does not require a Transformer decoder.
However, it is challenging to train a HuBERT encoder with a randomly initialized decoder with only 10h labelled data (rows 4 and 5).
In contrast, our \modelfull can easily adapt to an ASR task with only limited supervision (row 6).
In row 7, we show that it is more important to have a deeper encoder (8 layers) and a shallower decoder (4 layers), which allows closing the gap between CTC fine-tuning and Seq2Seq models.
In row 8, we use \modelfull as a second stage pre-training --- the encoder of \modelfull is initialized with a pre-trained HuBERT model. 
As we can see, with only a relatively small amount of additional pre-training cost, \modelfull is able to significantly improve HuBERT's Seq2Seq performance.
This demonstrates that HuBERT and \modelfull pre-training are complimentary.

\paragraphsq{Transducer Models}

We conduct similar experiments on a Transducer~\citep{graves2012sequence} model architecture and show that \modelfull pre-training is not restricted to Seq2Seq models (rows 11 and 12). Because Transducers require $O(mnV)$ space complexity ($m$, $n$, and $V$ are input sequence length, output sequence length, and output vocabulary size), we set $C = 25$ and $V = 1000$ to fit the model into a single GPU memory, which may hurt its performance compared to the Seq2Seq counterparts. We study the impact of  $C$ and $V$ in \shortautoref{subsec:more_ablation}.
We also reduce the number of decoder (a.k.a., label encoder in a transducer) layers because of memory issues. The pre-training learning rate is set to $5 \times 10^{-4}$ for Transducers.

\paragraphsq{Speech-to-text Transformer Models}

We show that \modelfull generalizes to other model architectures using the S2T Transformer model implemented by \citet{wang2020fairseqs2t}, which is a transformer-based Seq2Seq ASR model~\citep{mohamed2019transformers} that takes log-mel filterbank features as inputs and is widely used as a baseline under the fully supervised setup.
In this experiment, we closely follow the default hyperparameters in the codebase,\footnote{\url{https://github.com/pytorch/fairseq/blob/main/examples/speech_to_text/docs/librispeech_example.md}} except that we reduce the number of training iterations (300K originally) because the official example uses the complete 960h labelled data, while we use a semi-supervised setup with only 10h or 100h labelled data.
\shortautoref{tab:s2t-exp} shows the WERs of an S2T Transformer-small trained from scratch and a \modelfull pre-trained counterpart.
With merely 10h labelled data, it is impossible to train the S2T Transformer from scratch (row 1).
In contrast, with \modelfull pre-training, the model can learn meaningful patterns from limited labels (row 2).
With 100h labelled data, a model \modelfull pre-trained with 960h unlabelled data can significantly improve the WER of the model (32.8\% to 26.8\% on LibriSpeech test-other) while requiring merely one third of the fine-tuning time on the 100h ASR data (rows 3 and 4).

\begin{table}[t]
    \centering
    \tablestyle{3pt}{1.0}
    \scriptsize
    \begin{tabu}{clccccc}
    \toprule
    ASR Type & Pretraining method & PT Iter. & dev-o & test-o \\
    \midrule
    \multicolumn{4}{l}{\textit{10h labelled data:}} \\
    CTC & W2V2-base & 400K  & 17.4  & 17.6 \\
    CTC & HuBERT-base & 400K  & 16.9  & 17.2\\
    Seq2Seq & HuBERT-base & 400K & 27.2 & 28.2 \\
    \rowfont{\color{blue}}
    Seq2Seq & \modelsize{HuBERT-base} & 400K + 25K & 15.0 & 15.7 \\
    \rowfont{\color{blue}}
    Seq2Seq & \modelsize{HuBERT-base} & 400K + 100K & 14.7 & 15.3\\
    \midrule
    \multicolumn{4}{l}{\textit{10h labelled data + large model:}} \\
    CTC & W2V2-large & 400K  & 9.8 & 10.0\\
    CTC & HuBERT-large & 400K  & 9.9 & 10.2 \\
    Seq2Seq & HuBERT-large & 400K  & 17.4 & 18.7 \\
    \rowfont{\color{blue}}
    Seq2Seq & \modelsize{HuBERT-large} & 400K + 25K & 9.5 & 9.8 \\
    \midrule
    \multicolumn{5}{l}{\textit{100h labelled data:}} \\
    CTC & HuBERT-base & 400K & 13.6 & 12.9 \\
    \rowfont{\color{blue}}
    Seq2Seq  & \modelsize{HuBERT-base} & 400K + 100K  & 11.0 & 11.2 \\
    \bottomrule
    \end{tabu}
    \caption{Librispeech dev-other WER with different amount of labelled data. \Modelfull as a second stage pre-training method significantly improves the Seq2Seq model performance and allows it to match or even outperform CTC fine-tuning. No language models are used.}
    \label{tab:ls-10h-base}
\end{table}

\subsubsection{Standard Model Size Model Experiments}

We use the official HuBERT-base and HuBERT-large pre-trained on LibriSpeech and LibriLight as the initialization of \modelfull encoders following our observation that \modelfull works best as second stage pre-training. 
We further pre-train the models on the same corpus with relatively few updates (25K or 100K iterations compared to 400K interactions for HuBERT models). For the decoder, we use six Transformer blocks with the same width and number of heads as the encoder.
Our \modelsize{HuBERT-base} and \modelsize{HuBERT-large} models take 14 and 49 hours to be fine-tuned for 25K updates on eight NVIDIA V100 GPUs, a relatively small compute budget (less than 10\%) compared to training HuBERT models (usually 400K updates).

\shortautoref{tab:ls-10h-base} shows WER for standard sized models.
Even with a well trained HuBERT-base, fine-tuning with the Seq2Seq architecture using a randomly initialized decoder is inferior to the simple CTC objective (rows 2 and 3).
Using \modelfull closes the gap or even makes Seq2Seq models outperform CTC counterparts (row 4). However, we observe that pre-training longer does not help  for \modelsize{HuBERT-base} (row 5). The reason might be that the encoder has already been pre-trained. This shows that the second-stage pre-training can be stopped early without hurting performance.
We also compare models with 100h labelled data to confirm our observation.

\begin{table}[t]
    \tablestyle{3pt}{1.0}
    \scriptsize
    \centering
    \begin{tabu}{clcc}
        \toprule
        Model Type & PT method & dev & test \\
        \midrule
        Pipeline & W2V2-large + DeBERTa-large & 63.3 & 57.8 \\
        Pipeline & W2V2-large + LM + DeBERTa-large & 74.9 & 69.6 \\
        \midrule
        E2E CTC & W2V2-large & 55.6 & 50.9 \\
        E2E CTC & W2V2-large + LM & 70.2 & 64.8 \\
        E2E Seq2Seq & HuBERT-large & 64.0 & 58.5 \\
        \rowfont{\color{blue}}
        E2E Seq2Seq & \modelsize{HuBERT-large} & 71.7 & 65.4 \\
        \midrule
        \multicolumn{4}{l}{\textit{SoTA using external data:}} \\
        E2E CTC & W2V2-base + Distill-NLP (500h ASR data) & 82.2 & N/A \\
        \bottomrule
    \end{tabu}
    \caption{Development and test F1 scores (\%) on SLUE-VoxPopuli NER~\citep{shon2021slue}. \Modelfull achieves the best performance among all end-to-end methods without access to a language model. A pipeline model using an NLP pre-trained model (DeBERTa-large) remains the best among all the approaches. The numbers in the first four rows are provided by \citet{shon2021slue}. The last row is the state-of-the-art using external data and knowledge distillation from an NLP model, the test score is not available in their currently published paper~\citep{pasad2021use}. }
    \label{tab:slue-vp-ner}
\end{table}

\subsection{Spoken Named Entity Recognition (SNER)}
Spoken named entity recognition has recentely received significant attention~\citep{ghannay2018end,yadav2020end,shon2021slue}.
It is often addressed in one of two ways: end-to-end (E2E) or pipeline (i.e., a combination of an ASR model and an NLP NER model).
\citet{shon2021slue} show that the pipeline approach is the state-of-the art, while more recent work~\citep{pasad2021use} shows that with additional external data the order can be reverted.

End-to-end models are usually trained with a CTC objective~\citep{ghannay2018end,yadav2020end,shon2021slue}.
In our experiment, we explore the potential of applying Seq2Seq to SNER.
Following \citet{shon2021slue}, we add special tokens around named entities in the transcription and train the model in an ASR manner, which results in a model that detects named entities as it transcribes the audio inputs.

\shortautoref{tab:slue-vp-ner} shows F1 scores on SLUE-VoxPopuli dev and test sets.
While pipeline models remains the best without external data, Seq2Seq models significantly outperform the CTC baseline\footnote{\citet{shon2021slue} show that W2V2-base is slightly better (50.2\% vs. 49.8\% test F1) than HuBERT-base and only report the scores of W2V2-large, so we choose it as the CTC baseline.} without a language model (LM).
\modelsize{HuBERT-large} further improves performance (from 58.5\% to 65.4\% test F1), and obtains the best F1 score among all E2E models without external data, outperforming the best CTC model with an LM. 
Using external data~\citep{pasad2021use} remains the state-of-the-art. We leave the study of leveraging external data for Seq2Seqs model for future work.

\begin{table*}[t]
    \tablestyle{3pt}{1.0}
    \scriptsize
    \centering
    \begin{tabu}{l|cc|cccccccccccc|c}
    \toprule
    & \multicolumn{2}{c|}{Param (M)} & Tr & Ar & Et & Mn & Nl & Sv-SE & Lv & Sl & Ta & Ja & Id & Cy & Avg. \\
    Model & Enc & Dec & 4h & 2h & 3h & 3h & 7h & 2h & 2h & 2h & 2h & 2h & 2h & 2h & - \\
    \midrule
    HuBERT-large & 315 & 102 & 3.4 & 5.0 & 0.6 & 0.4 & 3.6 & 4.4 & 2.8 & 2.6 & 0.4 & 1.7 & 2.3 & 3.2 & 2.5 \\
    \rowfont{\color{blue}}
    \modelsize{HuBERT-large} & 315 & 102 & 3.7 & 5.1 & 0.7 & 0.2 & 6.5 & 4.9 & 3.6 & 3.9 & 0.0 & 0.9 & 3.5 & 3.3 & 3.0 \\
    XLS-R (0.3B) & 315 & 102 & 3.7 & 8.1 & 0.6 & 0.3 & 3.5 & 5.3 & 3.1 & 5.3 & 0.0 & 2.0 & 3.3 & 3.4 & 3.2 \\
    \rowfont{\color{blue}}
    \modelsize{XLS-R (0.3B)} & 315 & 102 & 4.5 & 10.5 & 2.4 & 0.3 & 12.2 & 8.8 & 4.8 & 5.9 & 0.0 & 1.9 & 5.0 & 5.7 & 5.2 \\
    \midrule
    \multicolumn{15}{l}{\textit{Use an mBART decoder (pretrained on multilingual machine translation data):}} \\
    XLS-R (0.3B) + mBART-ML50N1 & 315 & 459 & 4.6 & 3.0 & 3.5 & 0.4 & 22.0 & 10.3 & 6.0 & 6.6 & 0.2 & 0.6 & 1.4 & 2.5 & 5.1 \\
    XLSR-53 + mBART-ML50N1 & 315 & 459 & 3.7 & 1.2 & 0.7 & 0.6 & 20.5 & 2.8 & 1.9 & 0.5 & 0.1 & 0.4 & 0.6 & 5.6 & 3.2 \\
    VP-100K + mBART-ML50N1 & 315 & 459 & 0.9 & 0.8 & 4.6 & 0.3 & 18.3 & 11.7 & 9.0 & 8.1 & 0.1 & 0.2 & 0.7 & 0.6 & 4.6 \\
    XMEF-En + mBART-ML50N1 & 315 & 459 & 4.8 & 2.8 & 1.5 & 0.9 & 14.2 & 5.0 & 4.9 & 5.0 & 0.8 & 1.7 & 3.7 & 2.3 & 4.0 \\
    XMEF-X + mBART-ML50N1 & 315 & 459 & 9.4 & 6.4 & 2.5 & 1.2 & 24.0 & 4.0 & 5.0 & 5.6 & 0.9 & 1.0 & 2.8 & 8.1 & 5.9 \\
    \midrule
    \multicolumn{15}{l}{\textit{SoTA: Use a larger encoder + an mBART decoder (pretrained on multilingual machine translation data):}} \\
    XLS-R (2B) + mBART-ML50N1 & 2162 & 459 & 16.7 & 17.1 & 11.1 & 1.6 & 31.7 & 29.6 & 19.5 & 19.6 & 0.5 & 3.5 & 16.5 & 14.0 & 15.1 \\
    \bottomrule
    \end{tabu}
    \caption{Test BLEU scores on CoVoST 2 low-resource X-to-En language pairs. When limited labels are provided, using \modelfull as a second stage pre-training consistently improves the performance of HuBERT-large and XLS-R (0.3B). \modelsize{XLS-R (0.3B)} is on par with the XLS-R (0.3B) + mBART-ML50N1 which uses a large mBART decoder pre-trained on a 50-language text corpus and fine-tuned on 50 languages to English machine translation corpus with $4\times$ parameters. The scores in the second and third groups are from \citet{babu2021xls}. VP-100K is a wav2vec 2.0 large pre-trained on VoxPopuli 100K hour multilingual data~\citep{Wang2021VoxPopuliAL}, XLSR-53~\citep{Conneau2020UnsupervisedCR} is a wav2vec 2.0 pre-trained on 53-language audios, and XMEF-En and XMEF-X are efficient fine-tuning method proposed by \citet{li2020multilingual}. 
    }
    \label{tab:covost-X-en-small}
\end{table*}

\begin{table}[h]
    \centering
    \tablestyle{3pt}{1.0}
    \scriptsize
    \begin{tabu}{l|cccc|c}
    \toprule
    & Fr & De & Es & Ca & Avg. \\
    \midrule
    \rowfont{\color{blue}}
    \modelsize{HuBERT-large} & 33.2 & 26.9 & 34.0 & 29.7 & 31.0 \\
    \rowfont{\color{blue}}
    \modelsize{XLS-R (0.3B)} & 33.0 & 28.0 & 33.8 & 29.9 & 31.1 \\
    \midrule
    \multicolumn{6}{l}{\textit{Use an mBART decoder (pretrained on multilingual machine translation data):}} \\
    XLS-R (0.3B) + mBART-ML50N1 & 32.9 & 26.7 & 34.1 & 28.7 & 30.6 \\
    XLSR-53 + mBART-ML50N1 & 32.3 & 26.9 & 33.3 & 28.6 & 30.3 \\
    VP-100K + mBART-ML50N1 & 30.4 & 23.4 & 31.1 & 25.7 & 27.7 \\
    XMEF-En + mBART-ML50N1 & 35.0 & 28.2 & 35.2 & 31.1 & 32.4 \\
    XMEF-X + mBART-ML50N1 & 36.1 & 30.6 & 38.1 & 31.8 & 34.2 \\
    \midrule
    \multicolumn{6}{l}{\textit{SoTA: Use a large encoder and an mBART decoder:}} \\
    XLS-R (2B) + mBART-ML50N1 \hspace{40pt}     & 37.6 & 33.6 & 39.2 & 33.8 & 36.1 \\
    \bottomrule
    \end{tabu}
    \caption{Test BLEU scores on CoVoST-2 X-to-En high-resource language pairs. We compare with the same baselines as \shortautoref{tab:covost-X-en-small}.}
    \label{tab:covost-X-en}
\end{table}

\begin{table}[t]
    \tablestyle{3pt}{1.0}
    \scriptsize
    \centering
    \begin{tabu}{l|cccc|c}
    \toprule
    Model & De & Ca & Ar & Tr & Avg. \\
    \midrule
    HuBERT-large & 5.0 & 7.9 & 1.6 & 1.5 & 4.0 \\
    \rowfont{\color{blue}}
    \modelsize{HuBERT-large} & 9.7 & 13.0 & 3.1 & 2.9 & 7.2 \\
    \bottomrule
    \end{tabu}
    \caption{Test BLEU scores on CoVoST-2 En-to-X language pairs under a low-resource fine-tuning setup where only 10h labelled audio is provided.}
    \label{tab:covost-en-X-small}
\end{table}

\begin{table}[t]
    \centering
    \tablestyle{3pt}{1.0}
    \scriptsize
    \begin{tabu}{l|cccc|c}
    \toprule
     & De & Ca & Ar & Tr & Avg. \\
    \midrule
    W2V2-large ~\citep{Wang2021LargeScaleSA} & 23.8 & 34.0 & 18.0 & 15.4 & 22.8 \\
    \rowfont{\color{blue}}
    \modelsize{HuBERT-large} & 27.2 & 32.7 & 19.7 & 18.0 & 24.4 \\
    \midrule
    \multicolumn{6}{l}{\textit{Use an mBART decoder (pretrained on multilingual machine translation data)}} \\
    XLS-R (0.3B) + mBART-ML501N & 23.6 & 28.7 & 16.3 & 15.0 & 20.9 \\
    XLSR-53 + mBART-ML50N1 & 23.6 & 29 & 16.5 & 15.3 & 21.1 \\
    VP-100K + mBART-ML50N1 & 20.8 & 26.1 & 14.5 & 13.5 & 18.7 \\
    XMEF-JT + mBART-ML50N1 & 25.8 & 30.9 & 18.0 & 17.0 & 22.9 \\
    \midrule
    \multicolumn{6}{l}{\textit{Use a large encoder and an mBART decoder:}} \\
    W2V2 (0.72B) + mBART-ML501N & 27.0 & 32.7 & 19.4 & 17.7 & 24.2 \\
    XLS-R (1B) + mBART-ML501N & 26.1 & 32.1 & 19.2 & 17.1 & 23.6 \\
    XLS-R (2B) + mBART-ML501N & 28.3 & 34.2 & 20.7 & 18.6 & 25.5 \\
    \midrule
    \multicolumn{6}{l}{\textit{SoTA: Use self-training (labeling 60K hour of audio) + LM:}} \\
    W2V2-large + Self-training (labeling 60Kh data) & 26.5 & 34.1 & 20.2 & 17.5 & 24.6 \\
    \;\;\;\; + Decoding with LM  & 27.2 & 35.6 & 20.8 & 18.9 & 25.6 \\
    \bottomrule
    \end{tabu}

    \caption{
    Test BLEU scores on CoVoST-2 En-to-X language pairs. The complete 430h labelled audio data is used. Our \modelsize{HuBERT-large} outperforms XLS-R (1B)~\citep{babu2021xls} without using any multi-lingual data or using an mBART~\citep{liu2020multilingual} decoder pretrained on text data. It also reaches a similar performance by self-training~\citep{Wang2021LargeScaleSA} which is computationally expensive since it has to label 60K hours of audios and fine-tuning on them in a second iteration. We also compare with VP-100K~\citep{Wang2021VoxPopuliAL}, XLSR-53~\citep{Conneau2020UnsupervisedCR}, and XMEF-JT~\citep{li2020multilingual}.
    }
    \label{tab:covost-en-X}
\end{table}
\begin{table}[t]
    \tablestyle{3pt}{1.0}
    \scriptsize
    \centering
    \begin{tabu}{l|ccc}
        \toprule
        Model & 10h & 100h & 430h \\
        \midrule
        HuBERT-large & 5.0 & 20.0 & 26.5 \\
        \rowfont{\color{blue}}
        \modelsize{HuBERT-large} & 9.7 & 21.9 & 27.2 \\
        \midrule
        Improvement & 4.7 & 1.9 & 0.7 \\
        \bottomrule
    \end{tabu}
    \caption{Test BLUE scores on CoVoST-2 En-to-De with different amounts of labelled data. The gain from self-supervised pre-trained decoder decreases as the amount of the labelled data increases.}
    \label{tab:covost-en-de-scale}
\end{table}

\subsection{Speech-to-text Translation (ST)}

Encoder-decoder (i.e., Seq2Seq) models are considered particularly suitable for  speech-to-text translation tasks, where  the input and output sequences are not monotonically aligned. 

Our experiments extend beyond English. 
Besides the English \modelsize{HuBERT-large}, we also pre-train a \modelsize{XLS-R (0.3B)} which uses a multi-lingual XLS-R encoder~\citep{babu2021xls} pre-trained on 500K hours of audio in 128  languages. 
We use LibriLight for second stage pre-training and pre-train the model for only 25K updates.\footnote{25K updates is about 0.3 epoch of LibriLight, so the model only sees about 20K hour audio during second-stage pre-training.}

\paragraphsq{X-to-English Low-resource Experiments}
We experiment with 12 X-to-English language pairs where less than 10 hours of labelled data is available. We fine-tune both baselines and our models in a multi-task fine-tuning learning setup.
\shortautoref{tab:covost-X-en-small} shows test BLEU scores on each language pair.
\modelfull improves HuBERT and XLS-R (0.3B) models on most language pairs and achieves better average performance. Notably, our \modelsize{XLS-R (0.3B} can match the performance of XLS-R (0.3B) with an mBART-ML50N1 decoder which has 4$\times$ the number of parameters in the decoder, which is pre-trained on 50 language unlabelled texts and then fine-tuned on a 50-language-to-English machine translation corpus. Using \modelfull as a second-stage pre-training method makes the decoder size flexible and requires no additional text data.

\paragraphsq{X-to-English High-resource Experiments}
\shortautoref{tab:covost-X-en} shows the test BLEU scores of the four high-resource X-to-English   language pairs.
Our \modelsize{XLS-R (0.3B)} outperforms the counterpart using an mBART-ML50N1 decoder (row 1 vs. 3). 
Since the input audio is always in English, we do not observe performance gain with a multi-lingual pre-trained encoder (row 1 vs. 2).
XMEF models~\citep{li2020multilingual} and XLS-R (2B)~\citep{babu2021xls} achieve better BLEU scores with a mBART-ML50N1 decoder and more parameters.

\paragraphsq{English-to-X Low-resource Experiments}
Since English is a high resource language, all English-to-X language pairs in CoVoST-2 have 430 hours of labelled audios. We subsample a 10 hour subset for de-en, ca-en, ar-en, tr-en language pairs. We choose these four language pairs because they are studied more often in prior works~\citep{Wang2021LargeScaleSA,babu2021xls}.
\shortautoref{tab:covost-en-X-small} shows the test BLEU scores of the models fine-tuned on each  10h training set.
\modelsize{HuBERT-large} outperforms HuBERT-large baseline.

\paragraphsq{English-to-X High-resource Experiments}

\shortautoref{tab:covost-en-X} shows the test BLEU scores on four English-to-X pairs with the full 430h labelled data used for fine-tuning. Our \modelsize{HuBERT-large} model outperforms most of the models with the same size encoder and match the performance of the models with 2$\times$ or 3$\times$ encoder size (W2V2 (0.72B) and XLS-R (1B)). It also matches the performance of a wav2vec 2.0 large using 60K hours of audios for self-training.

\shortautoref{tab:covost-en-de-scale} shows the performance gain of \modelfull with different amounts of labelled data. A pre-trained decoder has higher impact on low resource setup, but the gain diminishes with more supervision.

\begin{table}[t]
    \centering
    \tablestyle{3pt}{1.0}
    \scriptsize
    \begin{tabular}{llcc}
        \toprule
        Method & Target Tokens & Length Compression & WER (\%) \\
        \midrule
        \modelfull & Psuedo Subwords & 17.4 \% & \textbf{38.1} \\
        \;\; - BPE tokenization & Pseudo Characters & 39.2 \% & 44.1 \\
        \;\;\;\; - Deduplication & Hidden Units & 100.0 \% & 96.5 \\
        \midrule
         & English Words & 5.2 \% &  \\
         & English Subwords & 8.4 \% &  \\
         & English Characters & 27.6 \% &  \\
        \bottomrule
    \end{tabular}
    \caption{WER (\%) on LibriSpeech dev-other set for pseudo ASR tasks with different target sequences. The length compression rate is the ratio between the decoder sequence length and the encoder sequence length (100\% means no compression) computed on dev-other. We also show the compression rate of English tokens. The compression rate of pseudo subwords lies between English subwords and English tokens.}
    \label{tab:ablation-pl}
    \vspace{-10pt}
\end{table}

\subsection{Ablation Study} \label{subsec:ablation}

We conduct ablation study using the small scale model in \shortautoref{subsubsec:small_model_exp} to understand the gain of each component in the generation of \methodfulls. 
\shortautoref{tab:ablation-pl} shows the results of pre-training with different type of target sequences.
Deduplication plays a vital role and using BPE tokenization to convert characters into subwords  further reduces WER.
We hypothesize that the duration of these characters is less relevant to the semantic of speech and it is more important to capture the transition of different characters.
We provide further ablation studies in \shortautoref{subsec:more_ablation}.

\section{Conclusion and Future Work}

We present \modelfull, a self-supervised learning framework for pre-training  speech processing models. 
Unlike existing methods, \modelfull pre-trains both encoder and decoder parameters, thereby allowing the two main components of common encoder-decoder architectures to benefit from pre-training. 
Critically, \modelfull requires raw audio data only. 
Instead of using aligned text, we create  a pseudo ASR task in which models transcribe audio inputs into \methodfull tokens.
We show that \modelfull closes the performance gap between encoder-decoder models and CTC models under low-resource conditions in ASR. 
On the SLUE-VoxPopuli spoken NER task, \modelfull achieves the best performance among all E2E models.
On speech-to-text translation tasks, \modelfull consistently improves the performance of HuBERT and XLS-R and rivals mBART decoder initialization, which requires additional language data, while being more flexible and having fewer parameters.

We demonstrate the potential of encoder-decoder models, which are applicable to more diverse tasks than CTC models. 
Understanding the impact of additional data (e.g., via knowledge distillation or self-training) or combining with external models (e.g., LMs trained on large text corpora) is an important direction for future studies.
Another interesting direction is to apply \methodfulls to generative spoken language models~\citep{lakhotia2021generative}.
Last but not least, it is possible to add CTC loss to the encoder during fine-tuning and do joint CTC/attention decoding~\citep{hori2017joint} to further improve the performance of \modelfull.

\paragraph{Limitations} We study ASR models that do not use language models (LMs). As seen in prior work~\citep{baevski2020wav2vec2,hsu2020hubert}, an external language model can boost the performance of CTC models.
Due to the architecture difference between CTC and Seq2Seq, different LMs have to be used with CTC and Seq2Seq models which further complicates the experimental setup. 
The order of the performance may be swapped when different LMs are applied.
Moreover, because CTC models are faster during inference, CTC models remain a valid choice for ASR if the performance difference is small.

{\small
\bibliography{refs}
\bibliographystyle{plainnat}
}

\clearpage

\appendix
\section{Additional Experiments}
\subsection{More Ablation Study} \label{subsec:more_ablation}
\shortautoref{tab:ablation-cv} shows the ablation study on the number of clusters and vocabulary size. As we can see, number of clusters has a large impact on the compression rate of the target sequence length. Number of clusters has a larger impact on the length compression rate and the WER after fine-tuned on LibriSpeech.
A moderate compression is preferable; too much or too less can hurt the performance.
Finally, \shortautoref{tab:ablation-k} shows the performance of using the optional average pooling. As we can see, having kernel size 2 is beneficial. The main goal for applying average pooling is to reduce prepossessing and pre-training cost: we store fewer HuBERT features and reduce the time spent on $k$-means and BPE tokenization for free.
\begin{table}[h]
    \tablestyle{3pt}{1.0}
    \scriptsize
    \centering
    \begin{tabular}{cccc}
    \toprule
    Number of clusters (C) & BPE vocab size (V) & Length Compression & WER (\%) \\
    \midrule
    25 & 1000 & 10.8 \% & 41.7 \\
    25 & 3000 & 9.2 \% & 42.5 \\
    25 & 10000 & 8.0 \% &  42.3 \\
    \midrule
    100 & 3000 & 14.5 \% & 39.1 \\
    100 & 10000 & 12.3 \% & 38.8 \\
    100 & 30000 & 10.8 \% & 38.9 \\
    \midrule
    500 & 3000 & 21.5 \% & 41.5 \\
    500 & 10000 & 17.4 \% & \textbf{38.1} \\
    500 & 30000 & 14.8 \% & 40.3 \\
    \bottomrule
    \end{tabular}
    \caption{Ablation study on number of clusters $C$ and BPE vocabulary size $V$. We fix average pooling size $K = 2$. We use small-scale model and report WER on LibriSpeech dev-other.}
    \label{tab:ablation-cv}
\end{table}
\begin{table}[h]
    \tablestyle{3pt}{1.0}
    \scriptsize
    \centering
    \begin{tabular}{ccc}
    \toprule
    Ave. Pool size (K) & Length Compression & WER (\%) \\
    \midrule
    1 & 18.5 \% & 41.1 \\
    2 & 14.8 \% & \textbf{40.3} \\
    4 & 10.7 \% & 41.3 \\
    \bottomrule
    \end{tabular}
    \caption{Ablation study on average pooling kernel size $K$. We use $C = 500, V = 30000$ in this experiment.}
    \label{tab:ablation-k}
\end{table}

\end{document}